\documentclass{article} 

\usepackage{hyperref}
\usepackage{url}
\usepackage[utf8x]{inputenc}
\usepackage{amsmath}
\usepackage{amsfonts}
\usepackage{graphicx}
\usepackage{amsmath}
\usepackage[colorinlistoftodos]{todonotes}
\usepackage{enumitem}
\newlist{arrowlist}{itemize}{1}
\setlist[arrowlist]{label=$\rightarrow$}

\usepackage{enumitem}
\newlist{pointlist}{itemize}{1}
\setlist[pointlist]{label=$\bullet$}

\usepackage[accepted]{icml2014}
\icmltitlerunning{Learning States Representations in POMDP}

\begin{document}

\twocolumn[
\icmltitle{Learning States Representations in POMDP}

\icmlauthor{Gabriella Contardo}{gabriella.contardo@lip6.fr}
\icmlauthor{Ludovic Denoyer}{ludovic.denoyer@lip6.fr}
\icmlauthor{Thierry Artieres}{thierry.artieres@lip6.fr}
\icmlauthor{Patrick Gallinari}{patrick.gallinari@lip6.fr}
\icmladdress{Sorbonne Universites, UPMC Univ Paris 06, UMR 7606, LIP6, F-75005, Paris, France \\
CNRS, UMR 7606,  LIP6, F-75005, Paris, France 
}


\vskip 0.3in
]
\begin{abstract}

We propose to deal with sequential processes where only partial observations are available by learning a latent representation space on which policies may be accurately learned.
\end{abstract}
\section{Introduction}  

	We consider a \textit{Markov decision process} (MDP) defined by possible states $s \in \mathcal{S}$, possible actions $a \in \mathcal{A}$, transition between states $P(s' | s,a)$ and reward function $r(s,a)$. Standard \textit{reinforcement learning} (RL) approaches make the assumption that the input provided to the model (i.e. the state of the system) contains enough information to learn an optimal policy (i.e a function $\pi(s) \in \mathcal{A}$ that chooses which action to take in a state in order to maximize the expected discounted reward).
   When facing \textit{approximated reinforcement learning} problems, the input consists in a feature vector - called observation - which is assumed to fully characterize the current state of the process, thus allowing for an optimal action choice. However, this assumption is unrealistic in real-life applications where the observation is only a partial view of the current state provided by limited sensors\footnote{This more general case is also known as \textit{partially observable Markov decision processes} or POMDP}. For example, it is the case in \textit{visual reinforcement learning} problems where the input is a camera-based picture of the environment, which can not provide second-order informations such as the moving speed of the different elements of the system. Without this information, a policy learned from such an observation will probably give low quality results.  
   
We propose to address this problem by switching from the original observation space to a latent representation space, which we expect to be more informative, and then learn policies in this new space.
The model operates in two steps: (i) First, it learns how to find good representations on a set of randomly collected trajectories. This unsupervised operation is used to learn the system only once, and may be used to tackle different tasks sharing the same dynamical process. (ii) The model then infers new representations for any new trajectory, these representations being then used for discovering an optimal policy for a particular reward function.

Our approach is transductive  in the sense that 
 whenever a new observation occurs, the system has to recompute all the previous representations so that it best matches the whole observation sequence. 
 Although this process could be expensive in terms of computation, we show how to perform fast Monte-Carlo simulations in the representation space resulting in a high-speed algorithm.
In comparison to common representation learning algorithms which directly compute the representation given the observation, our approach does the opposite, considering that a good representation is a representation from which the observation can be computed. 


\section{Model}
   	Let us denote $s_t$ the state of the process at step $t$ and $o_t$ the observation corresponding to this state. 
    An observation is a feature vector of size $m$: $o_t \in \mathbb{R}^m$. Note that the learning agent only accesses the observation and does not know in which exact state the  process is. The latent representation of a state $s_t$ will be denoted $z_t \in \mathbb{R}^n$. 
    Finding an optimal policy at the representation level, $\pi(z_t)$, can be made using standard Reinforcement Learning techniques.

Our model is based on two ideas: \textbf{(i)} The latent representation $z_t$ of a state $s_t$ should contain enough information to compute the corresponding observation $o_t$. We thus consider a decoder function $d_\theta : \mathbb{R}^n \rightarrow \mathbb{R}^m$ that aims at computing the observation given the representation of the current state. \textbf{(ii)} The representation $z_t$ of a state $s_t$ should contain information about the dynamics of the system, allowing to compute the representation of the next state. This is handled through the use of a dynamical function $m_\gamma : \mathbb{R}^n \times \mathcal{A} \rightarrow \mathbb{R}^n$ such that $m_\gamma(z_t,a_t)$ aims at computing $z_{t+1}$.

We address now the two steps of this approach: \textbf{unsupervised learning from randomly sampled trajectories} which will consist in learning the decoder and the dynamical function, and \textbf{inferring representations on new states} which will consist in finding the latent representation sequence from a new observation sequence. 
   
    \subsection{Unsupervised Learning}
    
Given a sequence of observations and actions $(o_1,a_1,...,o_t,a_t)$, we define the following loss function:
\begin{equation}
\begin{aligned}
L(\mathbf{z},\theta,\gamma) =& \sum\limits_t \Delta_{dec}(d_\theta(z_t),o_t) \\
&+ \sum\limits_t \Delta_{dyn}(m_\gamma(z_t,a_t),z_{t+1})
\end{aligned}
\end{equation}
where $\Delta_{dec}$ measures the quality of the decoder, $\Delta_{dyn}$ measures the quality of the dynamical model and $\mathbf{z}$ is the sequence of latent representations. The value of this loss function directly reflects the ability of $\mathbf{z},\theta$ and $\gamma$ to explain the observations.  Given a set of $Q$ trajectories, learning resumes to:
\begin{equation}
\mathbf{z}^*,\theta^*,\gamma^* = argmin \sum\limits_{q \in [1;Q]} L(\mathbf{z}^q,\theta,\gamma)
\label{eq1}
\end{equation}
where $\mathbf{z}^q$ is the sequence of representations computed for trajectory number $q$. 

Learning produces both the optimal decoder $d_{\theta^*}$  and the dynamical function $m_{\gamma^*}$, together with the representation sequences corresponding to the $Q$ trajectories.

\subsection{Inferring new representations}

Knowing $d_{\theta^*}$ and $m_{\gamma^*}$, the next question is to compute representations for new trajectories. Consider that at time $t$, given a sequence of observations $(o_1,a_1,...,o_t,a_t)$, the $t$ first representations $z_1$ to $z_t$ have already been computed. We propose two methods to compute the representation $z_{t+1}$:

\paragraph{Exact Inference} Given a new observation $o_{t+1}$, the first method consists in solving: 
\begin{equation}
\mathbf{z}^* = argmin_{z_1,..,z_{t+1}} L(\mathbf{z},\theta^*,\gamma^*)
\label{eq2}
\end{equation}
This produces the optimal representation $z_{t+1}$ while revising previously computed representations $z_1$ to $z_t$. 
This characteristic of our model can be intepreted as a \textit{thinking process} since it means that any new information gathered by the system will make it revise the whole representation sequence. The drawback of such an inference schema is its high complexity : finding a new representation may be slow and the optimization must be performed at each step.

\paragraph{Fast Inference} The second method consists in using the dynamical function to directly compute the next representation through $z_{t+1} = m_{\gamma^*}(z_t,a_t)$. In that case, the new representation can be produced directly from $z_t$ without requiring the observation $o_{t+1}$. The advantages are twofold. First the computation of $m_{\gamma^*}(z_t,a_t)$ is fast, which makes this inference method particularly adapted for processes where data acquisition is slow, such as moving robots for example. Moreover, the dynamical function can be used as a learned simulator which allows one to compute Monte-Carlo simulations directly in the latent representation  space, and thus to discover an optimal policy without needing to compute new real-world trajectories. 


\section{Experiments}

\begin{table}
\begin{center}
\small{
\begin{tabular}{|c|c|c||c|} \hline 
Input & Model &  Nb. Dim. & Perf. \\ \hline \hline
FO & FObs &  - &  \textbf{0.943} \\
PO & FObs &  - & 0.579\\ \hline \hline
PO & FLat &  2 & 0.86 \\ 
PO & FLat &  3 & 0.861 \\ 
PO & FLat &  5 & \textbf{0.912}\\ \hline \hline
PO & FDyn &  2 & 0.769 \\ 
PO & FDyn &  3 & 0.797  \\ 
PO & FDyn &  5 & 0.81  \\ \hline \hline
PO & FPar &  2 & 0.882\\ 
PO & FPar &  3 & 0.892 \\ 
PO & FPar &  5 & 0.91\\ \hline \hline
\end{tabular}
}
\end{center}
\caption{Average Reward obtained over 5 runs}
\label{ttab}
\end{table}
    We present preliminary results obtained on a classical toy example of the domain: \textit{mountain car} \cite{suttonRL}. This problem has been studied in many different articles with many different variants. In this article, we consider that: (i) A state $s=(x,\dot{x})$ is defined by the position of the car $x$ and its speed $\dot{x}$. (ii) The initial state $s_1$ is generated by uniformly sampling $x$ and $\dot{x}$ and following a random policy during 5 steps. (iii) Each trajectory has a limited size $T$ - in these exeriments $T=100$. (iv) When the car reaches the goal, or when the maximum trajectory size is reached, the episode stops. The reward function measures the average number of sucessful trajectories. We consider two settings: The \textbf{full observation (FO)} setting where the entire knowledge of the current state is given to the system i.e $o_t = (x_t,\dot{x}_t)$ and the \textbf{partial observation (PO)} setting where observations only contain the position of the car at time step $t$ i.e $o_t = (x_t)$.      

    In order to learn an optimal policy, we use the \textbf{RCPI} algorithm \cite{RCPI} using a linear classification model with a hinge-loss. At each iteration of RCPI, we sample $1000$ states and simulate the current policy using only $1$ trajectory per state. The base model we are using is a combination of an $L_2$ regularized linear decoder with a linear+hyperbolic-tangent dynamical model. The $\Delta$ losses are $L_1$ norms since using an $L_2$ norms gives lower performances.

We consider four models: The \textbf{From observation model (FObs.)} directly considers that the representation $z_t$ of a state $s_t$ is the observation i.e $z_t = o_t$. This corresponds to the classical Approximated Reinforcement Learning context.
  The \textbf{From  latent model (FLat.)} computes the $z_t$ value by minimizing the objective function for each new state. The \textbf{From dynamical model (FDyn.)} computes the representation of the initial state $z_1$ by minimizing the loss, using the 5-sized trajectory that generated $s_1$, and then use the fast inference method. At last, the \textbf{partial (FPar.)} model computes the representation of a state $z_t$ by randomly choosing between FDyn and FLat.

\begin{figure}
\begin{center}
                \includegraphics[width=0.7\linewidth]{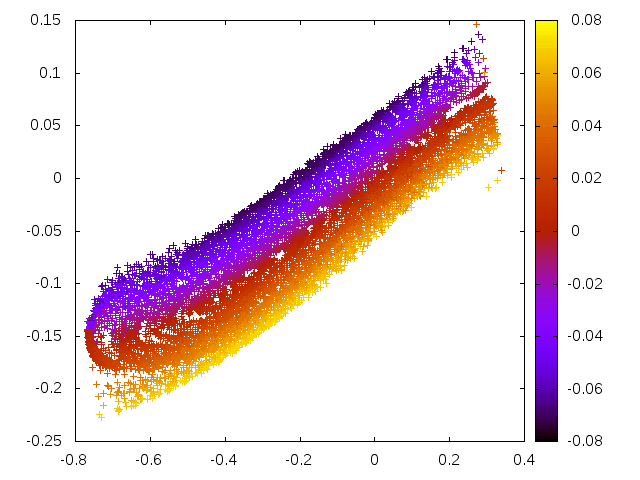}
\end{center}
               \caption{Plot of the 2D latent representations $z$ learned in the \textbf{PO} setting (the observation only includes $x$), the color indicates the speed value.}
\vspace{-0.5cm}
\label{fig1}               
               
\end{figure}

Experimental results are illustrated in Table \ref{ttab} which shows the performance - the expected reward - obtained by the policy found after 10 iterations of RCPI (averaged on 5 runs). The baseline corresponds to the first line where the full observation of the state is provided to the system (i.e speed and position). In that case, almost 95\% of the trajectories are sucessful. With our method, the best performance is obtained with the \textit{FLat} model in a latent space of size 5 - 91\% of success. Note that using alternative inference methods (FDyn and FPar) allows one to obtain good performance. Particularly, the FDyn results show the ability of the model to directly learn from the dynamical model, without acquiring observations and thus at a very high speed. An illustration of the learned latent space is given in Figure \ref{fig1}. As may be seen, states corresponding to different speeds are projected in different areas of the latent space, meaning the missing information has been recovered.

\section{Related work}

     Efficient approaches have been proposed to extract high-level representations using deep-learning \cite{bengio} but few studies have proposed extension to deal with sequential processes. A formal analysis has been proposed in \cite{Ryabko}. 
\\  Models concerning partially observable sequential processes have been proposed in the context of controlling tasks problems. For example, \cite{Schmidy90} and \cite{cuccu11} present models using recurrent neural networks (RNN) to learn a controller for a given task. In these approaches, informative representations are constructed by the RNN, but these representations are driven by the task to solve. Some unsupervised approaches have been recently proposed. In that case, a representation learning model is learned over the observations, without needing to define a reward function. The policy is learned afterward using these representations, by usually using classical RL algorithms. For instance, \cite{sRAAM} propose a model based on a recurrent auto-associative memory with history of arbitrary depth, while \cite{RNN} present an extension of RNN for unsupervised learning. \\
  In comparison to these models, our transductive approach is simultaenously based on unsupervised trajectories, and also allows us to choose which action to take even if observations are missing, by learning a dynamic model in the latent space.


     

\section{Conclusion}


We proposed a novel approach to 
learn representations on sequential processes when only partial observations are given. The model is unsupervised and transductive. 
It can be used for both inferring new representations, but also as a simulator, to predict what can happen in the future. Experiments on more realistic domains are currently under investigation.

\section*{Acknowledgements}
This work was performed within the Labex SMART supported by French state funds managed by the ANR within the Investissements d'Avenir programme under reference ANR-11-LABX-65 and by the Lampada project ANR-09-EMER-007.

\bibliography{ICLR2014}
\bibliographystyle{icml2014}
\end{document}